%% file: main.tex
\pdfoutput=1

\documentclass[11pt]{article}

\usepackage[final]{acl}
\usepackage{tabularx}
\usepackage{times}
\usepackage{latexsym}
\usepackage{amsmath}
\usepackage{amssymb}
\usepackage[T1]{fontenc}

\usepackage[utf8]{inputenc}

\usepackage{microtype}

\usepackage{inconsolata}

\usepackage{graphicx}
\usepackage{multirow}
\usepackage{colortbl}
\usepackage{makecell}
\usepackage{booktabs}
\usepackage{wrapfig}
\usepackage{graphicx}
\usepackage{ifthen}
\usepackage{caption}
\usepackage{subcaption}

\usepackage{makecell}
\usepackage{soul}
\usepackage{xspace}

\usepackage{pifont}
\newcommand{\xmark}{\ding{55}}%
\usepackage{colortbl}

\newcommand{\model}{{\textsc{FedKIM}\xspace}}

\title{{\model}: Adaptive Federated Knowledge Injection into Medical Foundation Models}

\author{Xiaochen Wang$^{1}$\thanks{Equal contribution.}, Jiaqi Wang$^{1}$\footnotemark[1], Houping Xiao$^2$, Jinghui Chen$^1$, Fenglong Ma$^1$\thanks{Corresponding author.}\\
$^1$Pennsylvania State University, $^2$Georgia State University\\
$^1$\texttt{\{xcwang, jqwang,jzc5917,fenglong\}@psu.edu},
$^2$\texttt{hxiao@gsu.edu}}

\begin{document}
\maketitle
\begin{abstract}
Foundation models have demonstrated remarkable capabilities in handling diverse modalities and tasks, outperforming conventional artificial intelligence (AI) approaches that are highly task-specific and modality-reliant. In the medical domain, however, the development of comprehensive foundation models is constrained by limited access to diverse modalities and stringent privacy regulations. To address these constraints, this study introduces a novel knowledge injection approach, {\model}, designed to scale the medical foundation model within a federated learning framework. {\model} leverages lightweight local models to extract healthcare knowledge from private data and integrates this knowledge into a centralized foundation model using a designed adaptive \textbf{M}ultitask \textbf{M}ultimodal \textbf{M}ixture \textbf{O}f \textbf{E}xperts (M$^3$OE) module. This method not only preserves privacy but also enhances the model's ability to handle complex medical tasks involving multiple modalities. Our extensive experiments across \textbf{twelve} tasks in \textbf{seven} modalities demonstrate the effectiveness of {\model} in various settings, highlighting its potential to scale medical foundation models without direct access to sensitive data. Source codes are available at ~\url{https://github.com/XiaochenWang-PSU/FedKIM}.
\end{abstract}

\input{section/intro_new}
\input{section/method_new}

\input{section/training_setup}

\input{section/evaluation_setup}
\input{section/ablation_study}

\input{section/related_work.tex}

\input{section/conclusion.tex}

\input{section/limitations}

\bibliography{anthology}

\input{section/appendix}

\end{document}

%% file: section/intro_new.tex
\section{Introduction}

Similar to large language models~\cite{zhao2023survey} and foundation models~\cite{zhou2023comprehensive}, medical foundation models~\cite{thirunavukarasu2023large,moor2023foundation} have achieved superior performance of handling diverse modalities and tasks within the medical domain. These models have the potential to revolutionize medical diagnostics and treatment by leveraging data-driven insights from large volumes of multimodal healthcare data. 
Due to the sensitive nature of medical data and the complexity of medical tasks, most existing medical foundation models usually rely on particular public medical datasets. This nature results in limitations of the existing medical foundation models, detailed as follows:



\begin{table}[t]
\centering
\caption{Summary of medical foundation models.}
\resizebox{0.49\textwidth}{!}{
\begin{tabular}{l|c|l}
\toprule
\textbf{Medical Foundation Model} & \textbf{Modalities}& \textbf{Tasks} \\ \hline
MMedLM2 ~\cite{qiu2024towards} & Text & Question-answering\\ \hline
LLava-Med\cite{li2023llava} & Text, Image & Visual Question-answering\\ \hline
Med-Flamingo\cite{moor2023med} & Text, Image & Visual Question-answering\\ \hline
PMC\_LLAMA\cite{wu2024pmc} & Text & Question-answering\\ \hline
BiomedGPT\cite{zhang2024generalist} & Text, Image & Visual Question-answering\\ \hline
BioMedLM\cite{bolton2024biomedlm} & Text & Question-answering\\ \hline
\multirow{5}{*}{GatorTron\cite{yang2022gatortron}} & \multirow{5}{*}{Text} & Clinical concept extraction\\ 
 & & Medical relation extraction\\ 
 & & Semantic textual similarity\\ 
 & & Natural language inference\\ 
 & & Question-answering\\ \hline
Med-PaLM\cite{singhal2023large} & Text & Question-answering\\ \hline
ChatDoctor\cite{li2023chatdoctor} & Text & Question-answering\\ 
\bottomrule
\end{tabular}}
\vspace{-0.1in}
\label{table:medical_models}
\end{table}

\begin{figure*}[t]
    \centering
    \includegraphics[width = 1\textwidth]{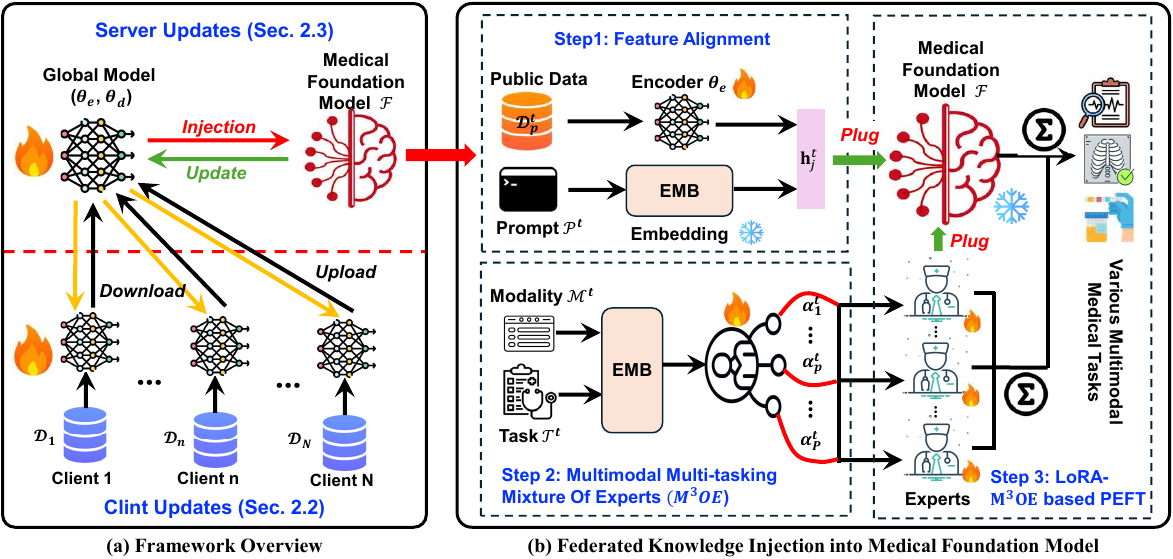}
    \vspace{-0.2in}
    \caption{Illustration of the proposed {\model}. (a) Framework overview, where the proposed {\model} contains client and server updates. (b) Federated knowledge injection, where {\model} first aggregates models uploaded from clients and then injects the aggregated model knowledge into medical foundation model $\mathcal{F}$ with three steps. ``PEFT'' in Step 3 denotes parameter-efficient fine-tuning.}
    \label{fig:overview}
\end{figure*}

(1) \textbf{Unrealistic to conduct large-scale centralized training}. 
The centralized training of medical foundation models presents significant challenges, primarily due to the difficulties in aggregating sensitive healthcare data. Regulations such as the Health Insurance Portability and Accountability Act (HIPAA) in the United States and the General Data Protection Regulation (GDPR) in the European Union impose strict privacy restrictions on the use of personal health information. This regulatory environment makes it impractical to collect and store large amounts of healthcare data in a single location, which is typically required for the effective training of high-performing medical foundation models.

(2) \textbf{Limited modality and task adaptability}. Current medical foundation models exhibit a high degree of specialization, constraining their effectiveness to a narrow range of downstream tasks within specific modalities, as outlined in Table~\ref{table:medical_models}. For instance, MMedLM~\cite{qiu2024towards} is tailored for text, while LLava-Med~\cite{liu2023llava} focuses on both image and text modalities. In practical settings, comprehensive medical decisions often require integrating multiple types of health data across various tasks. Yet, by being task or modality-specific, existing models fail to recognize and leverage the intricate relationships between different healthcare data modalities and tasks. 

The first limitation prevents training a medical foundation model from scratch in a centralized manner, while the second one exacerbates the challenge of developing a multimodal, multi-task medical foundation model. To overcome these obstacles, a viable solution is to scale existing medical foundation models and infuse them with medical knowledge. Given that medical data is stored on private clients, the federated learning (FL) paradigm~\cite{mcmahan2017communication,DBLP:journals/sensors/CheWZM23,DBLP:conf/cikm/Zhou0WH22} offers a promising approach in the medical domain~\cite{wang2022towards}, which is a decentralized and collaborative machine learning method where participants do not need to share data directly. Although several recent studies on federated foundation models~\cite{lu2023fedclip,chen2024feddat} have made progress, they primarily focus on enhancing services to local clients using existing foundation models. Importantly, none have specifically tackled the challenge of injecting novel medical knowledge into existing medical foundation models in a federated manner.



To tackle this new challenge, in this paper, we propose a novel approach: Federated Knowledge Injection for Medical foundation models ({\model}), as shown in Figure~\ref{fig:overview}. {\model} adopts a flexible design, allowing it to incorporate various types of medical modalities to handle a variety of medical tasks. Considering the real-world scenarios, {\model} deploys the medical foundation model only on the server side and leverages lightweight local models along with classic federated learning approaches to extract healthcare knowledge from private data. 

To effectively inject extracted medical knowledge into the foundation model, {\model} uses knowledge-rich parameters from the modality-specific encoders updated from the local end. To be specific, {\model} integrates this knowledge using parameter-efficient fine-tuning technique with a novel multitask multimodal mixture of expert module, namely M$^3$OE. M$^3$OE adaptively selects appropriate expert systems for handling specific tasks in given modalities, enabling {\model} to deal with tasks in complex medical contexts. 

Our experiments across 12 healthcare tasks with 7 modalities demonstrate the effectiveness of {\model}, providing a solid foundation for future exploratory research on the medical knowledge injection problem.


%% file: section/method_new.tex
\section{The proposed {\model} Framework}
\label{sec:method}

In this section, we first introduce the setup of the medical knowledge injection task (Section~\ref{sec:setup}). Next, we describe the proposed method, {\model}. As depicted in Figure~\ref{fig:overview}, {\model} consists of two main components: knowledge extractors (Section~\ref{sec:client}), which are deployed on local clients, and a knowledge injector (Section~\ref{sec:server}), which is deployed on the server.

\subsection{Framework Setups}\label{sec:setup}

\subsubsection{Client Setups} 
The goal of this work is to scale and enhance the predictive ability of medical large language models (LLMs) by incorporating medical knowledge from private client data in a federated manner. To achieve this, we employ $N$ clients, each representing a hospital or a medical institute holding private medical data $\mathcal{D}_n$. We assume that the private dataset $\mathcal{D}_n$ contains all medical modalities $\{\mathcal{M}_1, \cdots, \mathcal{M}_M\}$ and can perform all tasks $\{\mathcal{T}_1, \cdots, \mathcal{T}_T\}$. Each client trains a model $f_n = [\text{ENC}_n(); \text{DEC}_n()]$ using the data $\mathcal{D}_n$, where $\text{ENC}_n()$ is the set of multimodal encoders and $\text{DEC}_n()$ is the set of multi-task decoders/predictors. Thus, the model parameters $\boldsymbol{\theta}_n$ of $f_n$ can be divided into $\boldsymbol{\theta}_n^{enc}$ for the encoder and $\boldsymbol{\theta}_n^{dec}$ for the decoder, which will be further uploaded to the server.

\subsubsection{Server Setups} 
We deploy a generative medical foundation model on the server, denoted as $\mathcal{F}$. 
We aim to inject medical knowledge represented by $\{\boldsymbol{\theta}_1^{enc}, \cdots, \boldsymbol{\theta}_N^{enc}\}$ into $\mathcal{F}$ and simultaneously update $\{\boldsymbol{\theta}_1^{enc}, \cdots, \boldsymbol{\theta}_N^{enc}\}$ by absorbing new knowledge from $\mathcal{F}$. These updated encoders and the aggregated decoders will then be distributed to the corresponding clients for learning in the next communication round. To facilitate the updates of client parameters, we place a small amount of public data on the server, denoted as $\mathcal{D}_p$.


\subsection{Client Updates -- Knowledge Extraction from Private Clients}\label{sec:client}
This framework allows each client to handle $T$ tasks simultaneously. Although these tasks have different training data, the modalities are partially shared, which motivates us to design a simple client model with $M$ modality-specific encoders and $T$ task-specific decoders. Details of the encoders are listed in Appendix~\ref{sec:encoding}. 
We then use the following loss to train each client model:
\begin{gather}
\small
    \min_{\boldsymbol{\theta}_n} \mathcal{L}_n := \frac{1}{T}\sum_{t=1}^T \frac{1}{|\mathcal{D}_n^t|} \sum_{(\mathbf{x}_i^t, \mathbf{y}_i^t) \in \mathcal{D}_n^t} \ell^t(f_n(\mathbf{x}_i^t; \boldsymbol{\theta}_n), \mathbf{y}_i^t),\\
    \small
    f_n(\mathbf{x}_i; \boldsymbol{\theta}_n) = \text{DEC}_{n,t}(\text{ENC}_{n,m}(\mathbf{x}_i; \boldsymbol{\theta}_{n,m}^{enc}); \boldsymbol{\theta}_{n,t}^{dec}),
\end{gather}
where $\mathcal{D}_n^t$ is the task-specific dataset, $\mathbf{x}_i^t$ and $\mathbf{y}_i^t$ are the data features and the corresponding ground truths, and $\ell^t$ is the loss function for a specific task, such as cross-entropy.
$\text{ENC}_{n,m} \subseteq  \text{ENC}_n$ is the encoder for modality $\mathcal{M}_m$ with parameters $\boldsymbol{\theta}_{n,m}^{enc}$.
$\text{DEC}_{n,t} \subseteq  \text{DEC}_n$ is the decoder for the $t$-th task with parameters $\boldsymbol{\theta}_{n,t}^{dec}$. 
The number of modality-level encoders in $\text{ENC}_{n,m}$ is determined by the input data, while amount of tasks determines the number of task-oriented decoders. After the local training, we will upload the encoder and decoder parameters $\boldsymbol{\theta}_n^{enc}$ and $\boldsymbol{\theta}_n^{dec}$ to the server.

\subsection{Server Updates -- Knowledge Injection into Medical LLM}\label{sec:server}

\subsubsection{Knowledge Aggregation}
We assume that the predictive ability of $\mathcal{F}$ is better than the uploaded decoders $\{\boldsymbol{\theta}_1^{dec}, \cdots, \boldsymbol{\theta}_N^{dec}\}$, and useful knowledge is primarily contained in the encoders $\{\boldsymbol{\theta}_1^{enc}, \cdots, \boldsymbol{\theta}_N^{enc}\}$. Thus, on the server side, we aim to inject medical knowledge $\{\boldsymbol{\theta}_1^{enc}, \cdots, \boldsymbol{\theta}_N^{enc}\}$ into the LLM $\mathcal{F}$ with the help of public data $\mathcal{D}_p$. Before the injection, we first aggregate knowledge uploaded from each client in traditional federated learning manners such as FedAvg~\cite{mcmahan2017communication} or FedProx~\cite{li2020fedprox}, i.e., 
\begin{equation}
\begin{aligned}
    \boldsymbol{\theta}_e &= \textit{f}_{FL}([\boldsymbol{\theta}_e^1, \cdots, \boldsymbol{\theta}_e^M]), \\
    \boldsymbol{\theta}_d &= \textit{f}_{FL}([\boldsymbol{\theta}_d^1, \cdots, \boldsymbol{\theta}_d^M]),
\end{aligned}
\end{equation}
where $\textit{f}_{FL}$ can be flexibly replaced with any federated learning methods, such as \textit{personalized FL} methods~\cite{jiang2019improving,t2020personalized}, \textit{differential privacy-based FL} methods~\cite{hu2020personalized,el2022differential}, or \textit{adaptive FL} methods~\cite{reddi2020adaptive,wang2022communication,wang2022communication-compressed}. 


\subsubsection{Knowledge Injection}
\label{sec:knowledge_injection}
Effectively injecting medical knowledge $\boldsymbol{\theta}_e$ is challenging since the LLM $\mathcal{F}$ cannot directly use these diverse modality-specific encoders. To solve this challenge, we leverage a straightforward yet effective feature alignment strategy that follows the training of LLaVA~\cite{liu2024visual} by concatenating the modality embeddings with the task prompt. Subsequently, we embed our original \textbf{M}ultimodal \textbf{M}ulti-tasking \textbf{M}ixture \textbf{O}f \textbf{E}xperts (M$^3$OE) into the medical foundation model. M$^3$OE allows the medical foundation model $\mathcal{F}$ to adaptively select specific expert system given different combination of tasks and modalities. Next, we detail the process of knowledge injection.

\smallskip
\noindent\ul{\textbf{Step 1: Feature Alignment.}}
For each input data $(\mathbf{x}_j^t, \mathbf{y}_j^t) \in \mathcal{D}_p^t$ from the $t$-th task, we first obtain its feature representations using the aggregated encoders $\boldsymbol{\theta}_e$, i.e., $\mathbf{e}^t_j = [\mathbf{e}_j^1; \cdots; \mathbf{e}_j^M] = g(\boldsymbol{\theta}_e(\mathbf{x}_j^t)$), where $g(\cdot)$ is the linear mapping function.
We also embed the task prompt $\mathcal{P}^t$ using the encoder of $\mathcal{F}$, i.e., $\mathbf{p}^t = \text{EMB}_{\mathcal{F}}(\mathcal{P}^t)$., where $\text{EMB}_\mathcal{F}()$ is the text embedding layer of $\mathcal{F}$.
Then, the concatenation of the data feature $\mathbf{e}^t_j$ and the task prompt feature $\mathbf{p}^t$ will be used as the input of the encoder of $\mathcal{F}$, denoted as $\mathbf{h}_j^t = [\mathbf{e}^t_j; \mathbf{p}^t]$.

\smallskip
\noindent\ul{\textbf{Step 2: Multimodal Multi-tasking Mixture of Experts (M$^3$OE).}}
A naive solution is directly using the aligned feature $\mathbf{h}_j^t$ to generate the output. However, such a naive end-to-end fine-tuning approach not only has weak distinguishability of different tasks but also
ignores the generalization ability of {\model} to unseen tasks, even though the modalities have been encountered already. To address this issue, we develop a \textbf{M}ultimodal \textbf{M}ulti-tasking \textbf{M}ixture \textbf{O}f \textbf{E}xperts (M$^3$OE) module to allow {\model} to distinguish tasks dynamically. 

M$^3$OE takes both the task description $\mathcal{T}^t$ and the modality descriptions $\mathcal{M}^t$ associated with Task $\mathcal{T}^t$ as inputs to compute the relevance of each expert for the given task and modality, where $\mathcal{M}^t$ is the concatenation of descriptions of all modalities concerning Task $\mathcal{T}^t$. $\mathcal{T}^t$ and $\mathcal{M}^t$ are firstly encoded by the embedding layer of the foundation model $\mathcal{F}$, and subsequently processed to output weights for expert selection as follows:
\begin{equation}\label{eq:refined_attention}\small
\begin{aligned}
    \boldsymbol{\alpha}^t &= \text{softmax}\left(\text{MLP}\left(\text{Pooling}\left(\boldsymbol{\beta}^t\right)\right)\right), \\
    \boldsymbol{\beta}^t &= \frac{(\mathbf{W}_q \text{EMB}_\mathcal{F}(\mathcal{M}^t)) (\mathbf{W}_k \text{EMB}_\mathcal{F}(\mathcal{T}^t))^\top}{\sqrt{d_k}} \mathbf{W}_v \text{EMB}_\mathcal{F}(\mathcal{T}^t)
\end{aligned}
\end{equation}


where $\boldsymbol{\alpha}^t \in \mathbb{R}^P$ and $P$ is the number of experts. $\mathbf{W}_q$, $\mathbf{W}_k$, and $\mathbf{W}_v$ denote the attention matrices, and $d_k$ is the dimension size.

The proposed M$^3$OE effectively integrates the injected knowledge managed by two separate routers, resulting in a more streamlined and contextually aware computation of weights. The output, $\boldsymbol{\alpha}^t$, represents the attention-weighted selection of experts optimized for both the modality and the specific task. This approach provides the flexibility needed to handle complex medical scenarios by selecting the appropriate experts based on the context.

\smallskip
\noindent\ul{\textbf{Step 3: LoRA-M$^3$OE based Parameter-Efficient Fine-tuning.}} Finally, we generate the representation of each layer in $\mathcal{F}$ for the forward pass based on LoRA~\cite{hu2021lora} and the learned M$^3$OE weight using Eq.~\eqref{eq:refined_attention} as follows:
\begin{equation}
\mathbf{c}_j^t = \mathbf{W}_{\mathcal{F}} \mathbf{h}_j^t + \sum_{p =1}^P {\alpha}^t_p (\mathbf{B}_p \mathbf{A}_p \mathbf{h}_j^t),
\end{equation}
where $\mathbf{W}_{\mathcal{F}}$ denotes the frozen parameters of $\mathcal{F}$, 
$\mathbf{B}_p\mathbf{A}_p$ denotes the lower-rank adaptation module serving as the $p$-th expert system. We will fine-tune the proposed {\model} using the final output from $\mathcal{F}$ and the ground truth $\mathbf{y}_j^t$. 
The design balances efficacy and efficiency during knowledge injection, allowing {\model} to decently handle the complex nature of medical applications.

During the training, modality-specific encoders $\boldsymbol{\theta}_e$ gradually align with the medical LLM $\mathcal{F}$ that contains abundant knowledge acquired through pre-training. The alignment indicates prior knowledge in $\mathcal{F}$ is also extracted during injection, in the form of adjusted parameters restored in encoders. To benefit local models and boost knowledge injection in the next round, {\model} passes the updated encoders $\boldsymbol{\theta}_e$ and the aggregated decoders $\boldsymbol{\theta}_d$ back to local ends and performs the knowledge-driven iterative training until convergence.

%% file: section/training_setup.tex
\section{Experiment Setup}

\subsection{Task Introduction}\label{sec:training}
In this study, we have training tasks and validation tasks across different datasets and data modalities. To provide a clear illustration, we present them in Table~\ref{tab:tasks}.

\noindent\textbf{Training Task.} To examine the utility of the proposed {\model}, we leverage \textbf{four} classification tasks across \textbf{six} modalities to federatedly inject medical knowledge into the selected foundation model through multi-task training. Details regarding these tasks are available in Appendix~\ref{sec:training_tasks}. As emphasized in Section~\ref{sec:method}, we perform training on this suite of tasks in a multi-task pattern. 

\noindent\textbf{Validation Task.} Typical medical foundation models, such as MMedLM2~\cite{qiu2024towards}, often struggle with handling unseen tasks involving novel modalities. To evaluate the extent to which knowledge injection enables the medical foundation model to tackle unseen tasks, we compile \textbf{five} classification tasks (ECD, SP, PED, AD, and EBD) and \textbf{three} generation tasks (MR, SNC, and MS). Details on these tasks are provided in  Appendix~\ref{sec:validation_tasks}.

\begin{table*}[h!]
\centering
\caption{Tasks and modalities in this study.}
\label{tab:tasks}
\vspace{-0.15in}
\resizebox{1\textwidth}{!}
{
\begin{tabular}{c|l|c|c|c|c|c|c|c}
\cline{1-9}
\multirow{2}{*}{\textbf{Task Type}} & \multirow{2}{*}{\textbf{Task}} & \multicolumn{7}{c}{\textbf{Modality}} \\
\cline{3-9}
& & \textbf{Image} & \textbf{Signal} & \textbf{Vital signs} & \textbf{Lab events} & \textbf{Input} & \textbf{Output} & \textbf{Text} \\
\hline
\multirow{4}{*}{Training} & COVID-19 Detection (\textbf{CD}) & \cellcolor{red!20}{$\checkmark$} & \xmark & \xmark & \xmark & \xmark & \xmark & \xmark \\
\cline{2-9}
&  Lung Opacity Detection (\textbf{LOD}) & \cellcolor{red!20}{$\checkmark$} & \xmark & \xmark & \xmark & \xmark & \xmark & \xmark \\
\cline{2-9}
& ECG Abnormal Detection (\textbf{EAD}) & \xmark & \cellcolor{red!20}{$\checkmark$} & \xmark & \xmark & \xmark & \xmark & \xmark \\
\cline{2-9}
& Mortality Prediction (\textbf{MP}) & \xmark & \xmark & \cellcolor{red!20}{$\checkmark$} & \cellcolor{red!20}{$\checkmark$} & \cellcolor{red!20}{$\checkmark$} & \cellcolor{red!20}{$\checkmark$} & \xmark \\
\hline\hline
\multirow{8}{*}{\makecell[c]{Validation}} & Enlarged Cardiomediastinum Detection (\textbf{ECD}) & \cellcolor{green!20}{$\checkmark$} & \xmark & \xmark & \xmark & \xmark & \xmark & \xmark \\
\cline{2-9}
& Pleural Effusion Detection (\textbf{PED}) & \cellcolor{green!20}{$\checkmark$} & \xmark & \xmark & \xmark & \xmark & \xmark & \xmark \\
\cline{2-9}
& Atelectasis Detection (\textbf{AD}) & \cellcolor{green!20}{$\checkmark$} & \xmark & \xmark & \xmark & \xmark & \xmark & \xmark \\
\cline{2-9}
& Ectopic Beats Detection (\textbf{EBD}) & \xmark & \cellcolor{green!20}{$\checkmark$} & \xmark & \xmark & \xmark & \xmark & \xmark \\
\cline{2-9}
& Sepsis Prediction (\textbf{SP}) & \xmark & \xmark & \cellcolor{green!20}{$\checkmark$} & \cellcolor{green!20}{$\checkmark$} & \cellcolor{green!20}{$\checkmark$} & \cellcolor{green!20}{$\checkmark$} & \xmark \\
\cline{2-9}
& MedVQA-RAD (\textbf{MR}) & \cellcolor{green!20}{$\checkmark$} & \xmark & \xmark & \xmark & \xmark & \xmark & \cellcolor{green!20}{$\checkmark$} \\
\cline{2-9}
& MedVQA-Slake (\textbf{MS}) & \cellcolor{green!20}{$\checkmark$} & \xmark & \xmark & \xmark & \xmark & \xmark & \cellcolor{green!20}{$\checkmark$} \\\cline{2-9}
& Signal Noise Clarification (\textbf{SNC}) & \xmark & \cellcolor{green!20}{$\checkmark$} & \xmark & \xmark & \xmark & \xmark & \cellcolor{green!20}{$\checkmark$} \\
\hline

\end{tabular}
}
\vspace{-0.15in}
\end{table*}

\subsection{Data Partition}\label{sec:data}

For each training task, we divide the data into four parts in a ratio of 7:1:1:1. Specifically, 70\% of the data, $\mathcal{D}_{n}$, is private data evenly distributed to $N$ clients for training local models. Another 10\% of the data is public data, $\mathcal{D}_{p}$, placed on the server for tuning the foundation model. An additional 10\% of the data is development data, $\mathcal{D}_{d}$, kept on the server as a validation set. The remaining 10\% of the data, $\mathcal{D}_{t}$, is used as testing data for these tasks. More details regarding the data distribution can be found in Appendix ~\ref{sec:datasets}.

\subsection{Baselines}

Since the task of medical knowledge injection is novel and unexplored, there are no existing baselines. Therefore, we establish our own baselines, detailed as follows:

\noindent \textbf{FedPlug}. FedPlug acquires modality-specific encoders through the federated learning process described in Section \ref{sec:client}
. These encoders are then integrated into the foundation model for fine-tuning. By aligning multimodal medical input with the semantic space of the foundation model, FedPlug enables the model to handle multiple modalities. Throughout this process, only the aggregated encoders are trainable.

\noindent \textbf{FedPlug$_L$}. Building on the FedPlug framework, FedPlug$_L$ incorporates the Low-Rank Adaptation (LoRA) technique~\cite{hu2021lora} to better integrate multimodal features into the semantic space of the large language model (LLM), thereby optimizing the federated learning process. In addition to the trainable encoders in FedPlug, each layer of the LLM is equipped with a tunable LoRA module.

\subsection{FL Backbone Approaches}

We implement our {\model} based upon the following backbone approaches:

\noindent \textbf{FedAVG}~\cite{mcmahan2017communication} is a conventional federated learning method, producing a global model by aggregating distributed models $[\boldsymbol{\theta}^1, \cdots, \boldsymbol{\theta}^M]$ as follows:
$$\textit{f}_{avg}([\boldsymbol{\theta}^1, \cdots, \boldsymbol{\theta}^M]) = \frac{1}{N} \sum_{n=1}^N \boldsymbol{\theta}^n.$$

\noindent \textbf{FedProx}~\cite{li2020fedprox} aims to extend FedAvg by regularizing each local loss function with an $L_2$ term as follows:
\begin{equation}
    \min_{\mathbf{\boldsymbol{\theta}}^n} \mathcal{J}_n(\mathbf{\boldsymbol{\theta}}^n; \mathbf{\boldsymbol{\theta}}^{\ast}) = \mathcal{L}_n(\mathbf{\boldsymbol{\theta}}^n) + \frac{\lambda}{2}|| \mathbf{\boldsymbol{\theta}}^n- \mathbf{\boldsymbol{\theta}}^{\ast}||^2,
\end{equation}
where $\boldsymbol{\theta}^{\ast}$ is the global model, $\mathcal{L}_n(\cdot)$ is the corresponding loss function, and $\lambda$ is the hyperparameter for weighting.

\noindent \textbf{MMedLM-2}\footnote{\url{https://huggingface.co/Henrychur/MMedLM2}}~\cite{qiu2024towards} is an advanced unimodal Large Language Model. Benefiting from multilingual pre-training, MMedLM-2 achieves the state-of-the-art performance in multiple question answering tasks, thus selected as the backbone of our foundation model deployed on the server.

\begin{figure*}[t]
    \centering
    \begin{subfigure}[b]{0.45\textwidth}
        \centering
        \includegraphics[width=\textwidth]{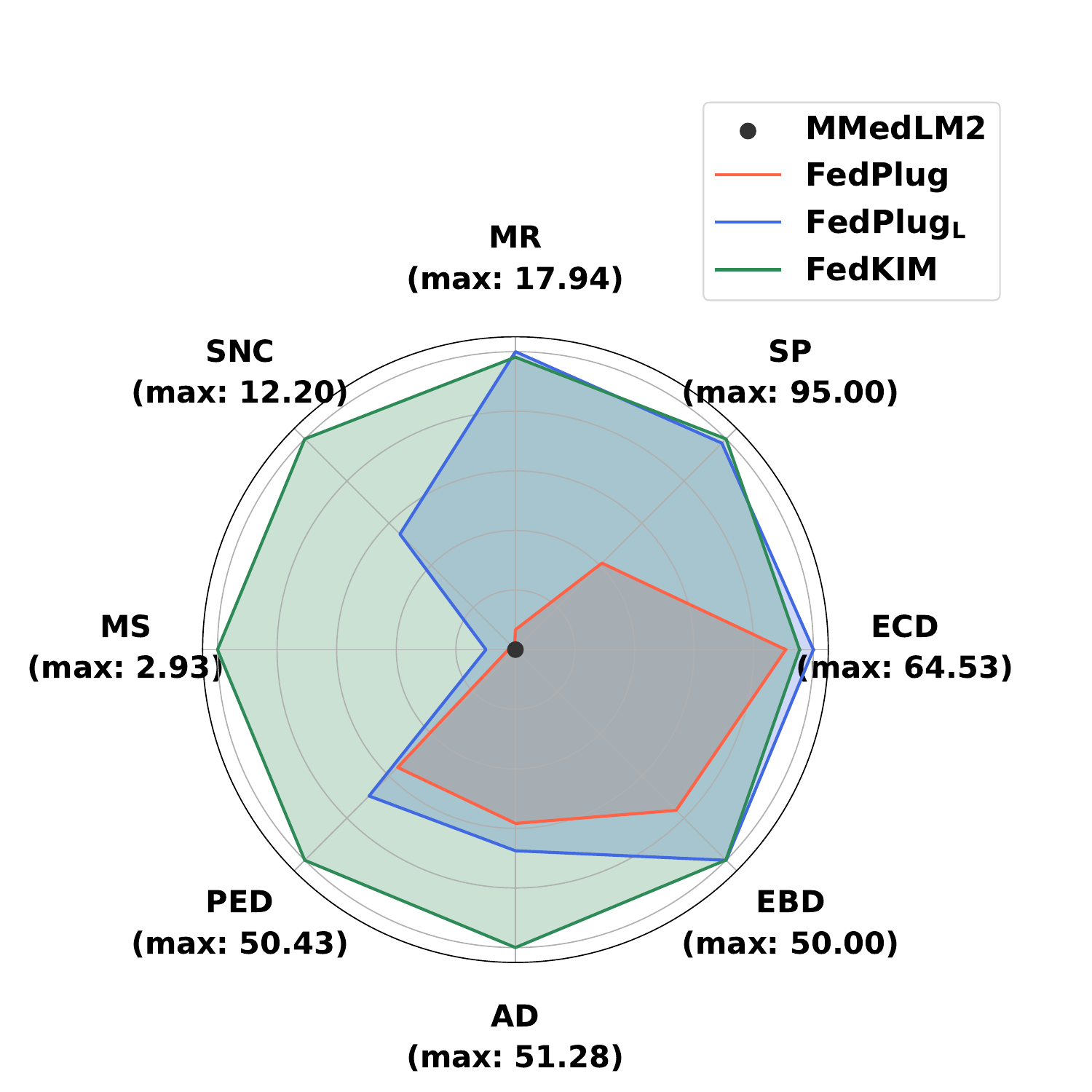}  
        \caption{FedAvg-based Knowledge Injection Performance.}
        \label{fig:zero_fedavg}
    \end{subfigure}
    \hfill
    \begin{subfigure}[b]{0.45\textwidth}
        \centering
        \includegraphics[width=\textwidth]{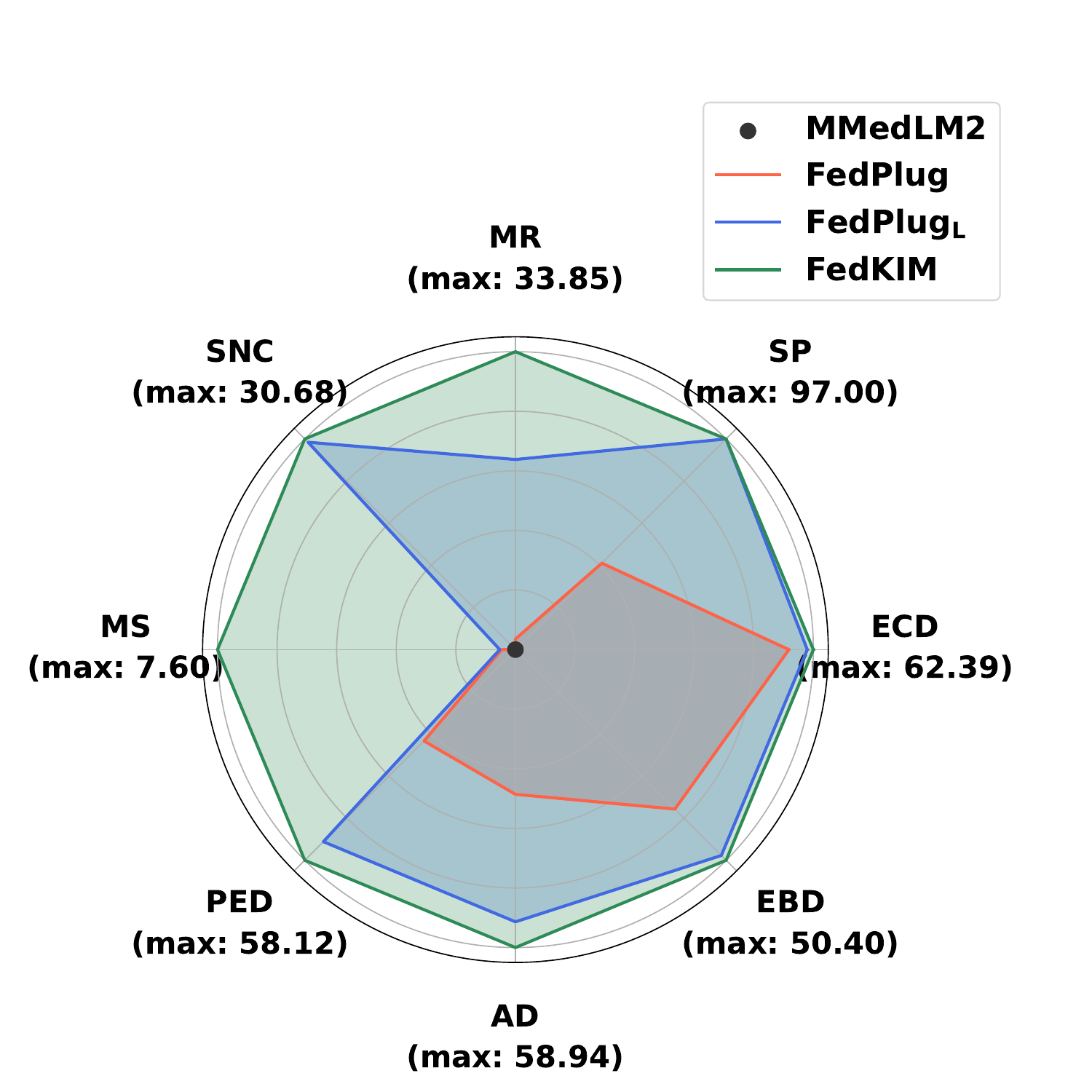}  
        \caption{FedProx-based Knowledge Injection Performance.}
        \label{fig:zero_fedprox}
    \end{subfigure}
    \vspace{-0.1in}
    \caption{Performance comparison between {\model} and baselines on the zero-shot evaluation.}
    \vspace{-0.15in}
    \label{fig:zero_shot}
\end{figure*}

\subsection{Implementation Details}

All experiments were conducted in an Ubuntu 20.04 environment using two NVIDIA A100 GPUs. We utilized MMedLM-2, the aforementioned state-of-the-art pre-trained medical language model, as the target of medical knowledge injection. The learning rate was set to \(5 \times 10^{-4}\) for the foundation model and \(1 \times 10^{-4}\) for the local models. $\lambda$ for FedProx was set to \(1 \times 10^{-4}\). Cross-entropy loss was used for training the local models, while the foundation model was optimized using general autoregressive loss. The number of clients \(N\) was set to 5, and the number of experts \(P\) was set to 12 for {\model}. To ensure a fair comparison, we set the number of communication rounds to 10 for all methods involved in the comparison.

%% file: section/evaluation_setup.tex
\section{Performance Evaluation}
\label{sec:evaluation}
We examine our proposed {\model}  from the zero-shot evaluation (subsection~\ref{sec:zeo-shot}) and fine-tuning evaluation (subsection~\ref{sec: fine-tuning}) perspectives.

\subsection{Zero-shot Evaluation}\label{sec:zeo-shot}
In the zero-shot evaluation, there is no overlap between the training tasks and evaluation tasks, which targets at examining the 
zero-shot capability of the medical foundation models enabled by \model. The experiment results on unseen tasks are shown in Figure~\ref{fig:zero_shot}, with FedAvg (Figure ~\ref{fig:zero_fedavg}) and FedProx (Figure ~\ref{fig:zero_fedprox}) as the backbone federated approaches. We use black { $\bullet$}, orange, blue, and green curves to denote MMedLM2, FedPlug, FedPlug$_L$, and {\model}, respectively. Accuracy for classification tasks and BLEU for generation tasks are used for visualization. Based on the experiment results, we provide the observations and discussion below: 

(1) The original foundation model MMedLM2 fails to do the zero-shot evaluation on the unseen tasks in the training process. This is due to its extremely limited multimodal capabilities.

(2) FedPlug, which only incorporates the federated encoder, performs the worst across all tasks, regardless of the type. This observation underscores the necessity of effectively utilizing public data to align the medical foundation model with external knowledge. Without proper integration, external knowledge—although derived through federated approaches on vast amounts of private data—cannot be directly assimilated into the medical foundation model.

(3) Even though FedPlug$_L$ approaches {\model}'s performance on several tasks, it still falls short, particularly in generation tasks like MedVQA. This indicates that the knowledge injected through FedPlug+LoRA does not fully generalize to unseen tasks, as training was exclusively performed on classification tasks. In contrast, {\model}, despite also being trained on classification tasks, achieves better performance on these tasks and maintains superior capability in handling unseen classification tasks. Comparing our {\model} with FedPlug$_L$, {\model} shows the superior performance on all the tasks, especially on the tasks of SNC ($\uparrow 82.36\%$ with FedAvg), PED ($\uparrow 43.92\%$ with FedAvg), and AD ($\uparrow 48.12\%$). On the other tasks, such as MR, EBD, and ECD, these approaches reach closed performance. This success is attributed to the M$^3$OE module, which enables {\model} to adaptively select appropriate experts to jointly handle novel tasks based on the context. Furthermore, our proposed {\model} works well with the federated backbones of FedAvg and FedProx. It also generally maintains the advantages of a more advanced federated learning method (FedProx) over the vanilla approach. Comparing Figure~\ref{fig:zero_fedavg} and Figure~\ref{fig:zero_fedprox}, the performance with FedProx generally outperforms the one with FedAvg on different tasks, such as PED $\uparrow 15.25\%$.

These observations further show the adaptability of {\model} to enable medical foundation models to have zero-shot capability across different tasks and federated learning frameworks.

\begin{table*}[t]
    \centering
    \caption{Fine-tuning evaluation for training tasks. {\xmark}  denotes incapacity.}
    \label{tab:multi-task}
    \vspace{-0.05in}
\resizebox{0.85\textwidth}{!}
{
    \begin{tabular}{c|l|c|ccc|ccc}
    \toprule
    \multirow{2}{*}{\textbf{Task}} & \textbf{Method}&\textbf{LLM}&\multicolumn{3}{c|}{\textbf{FedAvg}} & \multicolumn{3}{c}{\textbf{FedProx}}\\\cline{2-9}
    &\textbf{Metric}& \small{\textbf{MMedLM-2}}&\textbf{FedPlug} & \textbf{FedPlug$_L$} & \textbf{\model} & \textbf{FedPlug} & \textbf{FedPlug$_L$} & \textbf{\model} \\\hline
    \multirow{4}{*}{\makecell[c]{\textbf{Covid-19} \\ \textbf{Detection}}} &  \small{\textbf{Accuracy}}&\cellcolor{gray!10}\xmark&98.34&94.21&\cellcolor{blue!10}98.48 &86.11&95.73&\cellcolor{red!10}98.98\\
    & \small{\textbf{Precision}}&\cellcolor{gray!10}\xmark&96.07&99.27&\cellcolor{blue!10}96.61&65.09&99.32&\cellcolor{red!10}98.28\\
    &  \small{\textbf{Recall}} &\cellcolor{gray!10}\xmark&97.44&77.78&\cellcolor{blue!10}97.44&97.72&83.76&\cellcolor{red!10}97.72\\
    &  \small{\textbf{F1}} &\cellcolor{gray!10}\xmark&96.75&87.22&\cellcolor{blue!10}97.02&78.13&90.88&\cellcolor{red!10}98.00\\\hline
    \multirow{4}{*}{\makecell[c]{\textbf{Lung} \\ \textbf{Opacity} \\\textbf{Detection}}} &  \small{\textbf{Accuracy}}&\cellcolor{gray!10}\xmark&95.48&85.45&\cellcolor{blue!10}94.99&93.13&78.64&\cellcolor{red!10} 95.10\\
    & \small{\textbf{Precision}}&\cellcolor{gray!10}\xmark&98.22&99.85&\cellcolor{blue!10}98.32&93.74&72.20&\cellcolor{red!10}97.15\\
    & \small{\textbf{Recall}} &\cellcolor{gray!10}\xmark&92.95&74.03&\cellcolor{blue!10}91.90&92.95&95.58&\cellcolor{red!10}93.27\\
    & \small{\textbf{F1}} &\cellcolor{gray!10}\xmark&95.51&83.69&\cellcolor{blue!10}95.00&93.35&82.26&\cellcolor{red!10}95.17\\\hline
    \multirow{4}{*}{\makecell[c]{\textbf{ECG}\\ \textbf{Abnormal}\\ \textbf{Detection}}} &  \small{\textbf{Accuracy}}&\cellcolor{gray!10}\xmark&43.15&42.28&\cellcolor{blue!10}44.75&45.25&50.94&\cellcolor{red!10}58.46\\
    & \small{\textbf{Precision}} &\cellcolor{gray!10}\xmark&56.97&82.61&\cellcolor{blue!10}61.11&60.85&58.01&\cellcolor{red!10}58.46\\
    & \small{\textbf{Recall}} &\cellcolor{gray!10}\xmark&11.22&1.49&\cellcolor{blue!10}13.80&17.80&58.20&\cellcolor{red!10}100.00\\
    & \small{\textbf{F1}} &\cellcolor{gray!10}\xmark&18.74&2.93&\cellcolor{blue!10}22.52&27.55&58.10&\cellcolor{red!10}73.78\\\hline
    \multirow{4}{*}{\makecell[c]{\textbf{Mortality} \\ \textbf{Prediction}}} &  \small{\textbf{Accuracy}}&\cellcolor{gray!10}\xmark&84.11&53.63&\cellcolor{blue!10}90.01&82.41&91.42&\cellcolor{red!10}89.99\\
    & \small{\textbf{Precision}}&\cellcolor{gray!10}\xmark&16.35&10.96&\cellcolor{blue!10}35.88&13.87&47.57&\cellcolor{red!10}36.97\\
    & \small{\textbf{Recall}} &\cellcolor{gray!10}\xmark&21.43&63.04&\cellcolor{blue!10}23.29&16.64&15.22&\cellcolor{red!10}27.33\\
    & \small{\textbf{F1}} &\cellcolor{gray!10}\xmark&18.55&18.67&\cellcolor{blue!10}28.24&15.13&23.06&\cellcolor{red!10}31.43\\
    \bottomrule
    \end{tabular}
}   

\end{table*}

\subsection{Fine-tuning Evaluation}\label{sec: fine-tuning}

While injecting medical knowledge into foundation models demonstrates the potential for handling unseen tasks, it remains uncertain whether the enhanced foundation model can also perform well on previously encountered tasks. To address this, we conducted a fine-tuning evaluation, with the training process detailed in Section~\ref{sec:training} considered as fine-tuning for these tasks. The test sets for these tasks were used for evaluation, and the fine-tuning results are presented in Table~\ref{tab:multi-task}. For a comprehensive evaluation, we utilize accuracy, precision, recall, and F1 score as metrics for these tasks. 

Compared to the experiments on unseen tasks, it is evident that the knowledge-injected medical foundation model performs significantly better on familiar tasks. This showcases the explicit utilization of knowledge acquired through federated training. Similar to the zero-shot evaluation, approaches combined with FedProx consistently outperform those with FedAvg, underscoring the importance of effective knowledge extraction during the injection process.

Furthermore, {\model} consistently outperforms the two baselines, FedPlug and FedPlug$_L$. This competitive performance validates the design and effectiveness of the M$^3$OE module.

%% file: section/ablation_study.tex
\subsection{Ablation Study}

We conduct ablation studies on the COVID-19 detection and enlarged cardiomediastinum detection tasks to assess the impact of each module within our proposed \model in both fine-tuning and zero-shot settings. Retaining all other modules as in the main experiments, we explore the following variant settings: (1) {\model}$^{pub}$: Instead of utilizing knowledge from private datasets $\mathcal{D}_{n}$, this configuration solely leverages public dataset $\mathcal{D}_{p}$ for centralized training. Consequently, the federated training module discussed in Section~\ref{sec:client} is excluded, with the encoder $\boldsymbol{\theta}_e$ updated exclusively through public training as detailed in Section~\ref{sec:server}. (2) {\model}$^{\mathcal{T}}$: This variant omits the task description module that guides the expert selection process, testing the importance of task-specific information in routing the mixture of experts. (3) {\model}$^{\mathcal{M}}$: Similarly, we remove the modality description module to examine its influence on expert selection.

\begin{table}[t]
    \centering
    \caption{Ablation study results in the fine-tuning setting.}
    \label{tab:ablation_study_ft}
    \vspace{-0.05in}
\resizebox{0.9\columnwidth}{!}
{
 \begin{tabular}{l|cccc}
    \toprule
   \textbf{Setting} & \textbf{Accuracy} & \textbf{Precision} & \textbf{Recall} & \textbf{F1} \\\hline

    \textbf{\model$^{pub}$} &94.07 &90.63 &85.47 &87.98 \\
    \textbf{\model$^{\mathcal{T}}$} &98.34 &96.28 &97.12 &96.70 \\
    \textbf{\model$^{\mathcal{M}}$} &98.41 &97.13 &96.58 &96.86 \\
    \cellcolor{blue!20}\textbf{\model} & \cellcolor{blue!20}98.48 & \cellcolor{blue!20}96.61 & \cellcolor{blue!20}97.44 & \cellcolor{blue!20}97.02 \\
    \bottomrule
    \end{tabular}
}
\vspace{-0.1in}
\end{table}

\begin{table}[t]
    \centering
    \caption{Ablation study results in the zero-shot setting.}
    \label{tab:ablation_study_zs}
\resizebox{0.9\columnwidth}{!}
{
 \begin{tabular}{l|cccc}
    \toprule
   \textbf{Setting} & \textbf{Accuracy} & \textbf{Precision} & \textbf{Recall} & \textbf{F1} \\\hline

    \textbf{\model$^{pub}$} &59.82	&54.43	&84.40	&66.19 \\
    \textbf{\model$^{\mathcal{T}}$} &61.11	&55.80	&79.82	&65.66 \\
    \textbf{\model$^{\mathcal{M}}$} &60.26	&66.00	&30.28	&41.51 \\
    \cellcolor{blue!20}\textbf{\model} & \cellcolor{blue!20}61.54 & \cellcolor{blue!20}55.13 & \cellcolor{blue!20}93.58 & \cellcolor{blue!20}69.39 \\
    \bottomrule
    \end{tabular}
}
\vspace{-0.2in}
\end{table}

The results of the ablation studies are presented in Table~\ref{tab:ablation_study_ft} and Table~\ref{tab:ablation_study_zs}. They indicate that each component significantly enhances \model's performance. Specifically, a substantial decline in performance with {\model}$^{pub}$ highlights the crucial role of knowledge injected from local clients through federated learning. This locally enriched encoder allows the medical foundation model to better adjust to unseen modalities, thereby enhancing its effectiveness compared to models trained without this knowledge. Moreover, the absence of task or modality descriptions diminishes \model’s ability to manage specific tasks through multi-task training, validating the design of the M$^3$OE module. This module equips {\model} to effectively navigate complex healthcare scenarios that involve diverse tasks and modalities. In summary, the synergistic integration of local knowledge, along with the task and modality description modules, crucially bolsters the performance of our proposed \model.

%% file: section/related_work.tex
\section{Related Work}

\textbf{Medical Foundation Models.}  
Foundation models, known for their vast parameters and training datasets, have demonstrated impressive capabilities across domains~\cite{touvron2023llama,zhou2023comprehensive,yang2024foundation,li2024multimodal,abbasian2024foundation}, and are becoming increasingly prevalent in healthcare. Thirunavukarasu et al.~\cite{thirunavukarasu2023large} highlight the potential of large language models (LLMs) in clinical settings. Moor et al.~\cite{moor2023foundation} propose a generalist medical AI for diverse tasks using multimodal data. Specialized medical foundation models have been developed for disease detection~\cite{zhou2023foundation}, cancer biomarker identification~\cite{pai2024foundation}, echocardiogram interpretation~\cite{christensen2024vision}, image segmentation~\cite{zhang2023input}, and precision oncology~\cite{truhn2024large}. Despite these advancements, this area remains relatively unexplored compared to the general domain~\cite{wang2024recent}, primarily due to the complexity inherent in healthcare data~\cite{wang2023hierarchical, wang2024unity}.

\noindent \textbf{Federated Fine-tuning with Foundation Models.}  
Fine-tuning foundation models (FMs) with task-specific data is essential for improved performance in specialized tasks. Federated Learning (FL) supports this by utilizing locally stored data and distributed computational resources. Research in this field includes full tuning~\cite{deng2023mutual,fan2023fate}, partial tuning~\cite{peng2024fedpft,marchisio2022mini,khalid2023cefhri}, and parameter-efficient fine-tuning (PEFT)~\cite{lu2023fedclip,zhang2023fedpetuning}. Notably, \cite{lu2023fedclip} involves clients hosting FMs and exchanging adapters with the server, which aggregates and redistributes them. Similarly, FedPETuning~\cite{zhang2023fedpetuning} shares parts of client models for pre-trained language models in FL. Unlike these studies, which require clients to have FMs, our approach positions the medical FM on the server, facilitating collaborative enhancement of medical FM models without accessing local data.

\noindent \textbf{Parameter-efficient Fine-tuning on Foundation Model} Full-parameter Fine-Tuning of foundation models, while promising in terms of performance enhancement, requires extremely extensive computational resources. Consequently, researchers have investigated Parameter-efficient Fine-tuning (PEFT) techniques. PEFT methods aim to adapt pre-trained models to specific tasks using a minimal number of additional parameters. Low-Rank Adaptation (LoRA)~\cite{hu2021lora}, a widely recognized PEFT method, reduces the number of trainable parameters by factorizing weight matrices into low-rank representations, achieving significant parameter efficiency. Additionally, previous studies have utilized modular approaches, such as adapters~\cite{gao2023llama} and the Perceiver Resampler~\cite{alayrac2022flamingo}, to adapt new modalities to foundation models.

Researchers have explored combining the Mixture of Experts~\cite{jacobs1991adaptive} concept with Low-Rank Adaptation (LoRA) for Parameter-efficient Fine-tuning (PEFT)~\cite{li2024mixlora, wu2023mole}. To guide the selection of experts in complex scenarios, they have leveraged modality information~\cite{luo2024cheap, li2024uni}, instructions~\cite{chen2023octavius, chen2024llava, wu2023mole, li2024mixlora}, or pre-defined task IDs~\cite{liu2023moelora}. However, these MOE methodologies do not specifically address the complex, modality-diverse scenarios found in the healthcare domain.

%% file: section/conclusion.tex
\section{Conclusion}

This work introduces the concept of knowledge injection into medical foundation models, emphasizing its critical role and potential in the development of comprehensive medical models. We propose a novel approach, {\model}, designed to extract and inject healthcare knowledge into foundation models, thereby enhancing their ability to handle multiple tasks and modalities. {\model} leverages flexible federated learning techniques to extract knowledge from distributed medical data. The extracted knowledge is then injected into the foundation model using our proposed adaptive M$^3$OE module. Our exhaustive experimental results on 12 tasks and 7 modalities demonstrate the effectiveness of {\model} in diverse settings, showcasing its excellent capability in handling either encountered or unseen healthcare tasks. This study validates the potential of injecting knowledge into foundation models using federated learning, providing a crucial solution for developing a healthcare foundation model without accessing sensitive data.

%% file: section/limitations.tex
\section{Limitations}

This work explores the problem of medical knowledge injection within the PEFT framework. Due to current computational limitations, we have not yet combined Full-parameter Fine-Tuning with our proposed {\model}. Additionally, our study utilizes MMedLM2, which has 7 billion parameters, but injecting knowledge into larger foundation models is restricted by available computational resources. In future research, we plan to investigate the integration of knowledge injection with Full-parameter Fine-Tuning. We also aim to evaluate the efficacy of our approach on larger medical foundation models to further validate its scalability and potential.

\section*{Acknowledgements}
This work is partially supported by the National Science Foundation under Grant No. 2238275 and 2348541 and the National Institutes of Health under Grant No. R01AG077016.

%% file: section/appendix.tex
\appendix
\begin{table*}[t]
\centering
\caption{Details about the datasets.}
\label{tab:datasets}
\resizebox{1\textwidth}{!}
{
\begin{tabular}{c|l|c|c|c|c|c}
\toprule
\textbf{Task Type} & \textbf{Task}    & \textbf{Total Samples} & \makecell[c]{\textbf{Private}\\{Clients}} & \makecell[c]{\textbf{Public}\\(Server)} & \makecell[c]{\textbf{Development}} & \makecell[c]{\textbf{Testing}} \\ \hline
\multirow{4}{*}{\makecell[c]{Training \\Tasks}}
& Lung Opacity Detection            & 18,406                    &  12,880                   & 1,849                     &  1,841                   & 1,836                   \\ \cline{2-7}

& COVID-19 Detection              &  13,808                  &  9,665                  & 1,380                     &  1,380                  & 1,383                  \\ \cline{2-7}

& ECG Abnormal Detection               & 21,797                    &  15,259                    & 2,179                     &  2,180                   & 2,179                   \\ \cline{2-7}

& Mortality Prediction           & 38,129                    & 26,690                     & 3,812                     & 3,812                    & 3,813                   \\ \hline\hline

\multirow{8}{*}{\makecell[c]{Validation}}

& Enlarged Cardiomediastinum Detection           & 234                   &  \xmark                  & \xmark                      &  \xmark                   & 234 \\\cline{2-7}
& Pleural Effusion Detection            & 234                   &  \xmark                  & \xmark                      &  \xmark                   & 234 \\\cline{2-7}
& Atelectasis Detection           & 234                   &  \xmark                  & \xmark                      &  \xmark                   & 234 \\\cline{2-7}
& Sepsis Prediction           & 1,000                   &  \xmark                  & \xmark                      &  \xmark                   & 1,000 \\\cline{2-7}

& MedVQA-RAD            & 1,000                   &  \xmark                  & \xmark                      &  \xmark                   & 1,000                   \\ \cline{2-7}
& MedVQA-Slake            & 1,000                   &  \xmark                  & \xmark                      &  \xmark                   & 1,000                   \\ \cline{2-7}
& Signal Noise Clarification       & 1,000                    & \xmark                     & \xmark                      &  \xmark                   & 1,000                   \\ \cline{2-7}
& Ectopic Beats Detection       & 2,000                    & \xmark                     & \xmark                      &  \xmark                   & 2,000                   \\ \bottomrule

\end{tabular}
}
\end{table*}

\begin{table*}[t]
    \centering
    \caption{Single task fine-tuning evaluation.}
    \label{tab:single_task}
\resizebox{1\textwidth}{!}
{
    \begin{tabular}{c|l|c|ccc|ccc}
    \toprule
    \multirow{2}{*}{\textbf{Task}} & \textbf{Method} & \textbf{LLM} & \multicolumn{3}{c|}{\textbf{FedAvg}} & \multicolumn{3}{c}{\textbf{FedProx}}\\\cline{3-9}
    & \textbf{Metric} &\textbf{MMedLM-2} & \textbf{FedPlug} & \textbf{FedPlug$_L$} & \textbf{\model} & \textbf{FedPlug} & \textbf{FedPlug$_L$} & \textbf{\model} \\\hline

    \multirow{4}{*}{\makecell[c]{\textbf{COVID-19} \\ \textbf{Detection}}} 
    & \textbf{Accuracy} & \xmark & 92.34 & 99.56 & \cellcolor{blue!10}99.57 & 84.16 & 95.66 & \cellcolor{red!10}98.91 \\
    & \textbf{Precision} & \xmark & 93.59 & 98.87 & \cellcolor{blue!10}99.15 & 77.27 & 84.18 & \cellcolor{red!10}100.00 \\
    & \textbf{Recall} & \xmark & 74.92 & 99.43 & \cellcolor{blue!10}99.15 & 53.27 & 98.68 & \cellcolor{red!10}95.73 \\
    & \textbf{F1} & \xmark & 79.15 & 99.14 & \cellcolor{blue!10}99.15 & 63.07 & 90.85 & \cellcolor{red!10}97.82 \\\hline

    \multirow{4}{*}{\makecell[c]{\textbf{Lung} \\ \textbf{Opacity} \\ \textbf{Detection}}} 
    & \textbf{Accuracy} & \xmark & 89.42 & 51.69 & \cellcolor{blue!10}94.93 & 91.23 & 90.90 & \cellcolor{red!10}93.63 \\
    & \textbf{Precision} & \xmark & 84.69 & 51.74 & \cellcolor{blue!10}93.60 & 87.76 & 88.06 & \cellcolor{red!10}92.55 \\
    & \textbf{Recall} & \xmark & 97.16 & 99.79 & \cellcolor{blue!10}96.85 & 96.52 & 95.37 & \cellcolor{red!10}95.37 \\
    & \textbf{F1} & \xmark & 90.50 & 68.10 & \cellcolor{blue!10}95.19 & 91.93 & 91.57 & \cellcolor{red!10}93.94 \\\hline

    \multirow{4}{*}{\makecell[c]{\textbf{ECG} \\ \textbf{Abnormal} \\ \textbf{Detection}}} 
    & \textbf{Accuracy} & \xmark & 43.15 & 48.79 & \cellcolor{blue!10}58.46 & 45.25 & 50.80 & \cellcolor{red!10}58.00 \\
    & \textbf{Precision} & \xmark & 56.97 & 65.02 & \cellcolor{blue!10}58.46 & 60.85 & 58.47 & \cellcolor{red!10}58.34 \\
    & \textbf{Recall} & \xmark & 11.22 & 26.82 & \cellcolor{blue!10}100.00 & 17.80 & 54.67 &\cellcolor{red!10} 98.51 \\
    & \textbf{F1} & \xmark & 18.74 & 37.98 & \cellcolor{blue!10}73.78 & 27.55 & 56.51 & \cellcolor{red!10}73.28 \\\hline

    \multirow{4}{*}{\makecell[c]{\textbf{Mortality} \\ \textbf{Prediction}}} 
    & \textbf{Accuracy} & \xmark & 84.11 & 91.67 & \cellcolor{blue!10}64.16 & 82.41 & 90.98 & \cellcolor{red!10}57.38 \\
    & \textbf{Precision} & \xmark & 16.35 & 43.24 & \cellcolor{blue!10}12.86 & 13.87 & 40.68 & \cellcolor{red!10}13.25 \\
    & \textbf{Recall} & \xmark & 21.43 & 14.91 & \cellcolor{blue!10}56.21 & 16.64 & 14.91 & \cellcolor{red!10}72.98 \\
    & \textbf{F1} & \xmark & 18.55 & 22.18 & \cellcolor{blue!10}20.94 & 15.13 & 21.82 & \cellcolor{red!10}22.42 \\

    \bottomrule
    \end{tabular}
}
\end{table*}

\section{Details of Training Tasks}
\label{sec:training_tasks}

\noindent \textbf{COVID-19 Detection (CD)} involves identifying COVID-19 symptoms from X-ray images using the COVQU dataset ~\cite{rahman2021exploring} to evaluate the model's ability to interpret medical images.

\noindent \textbf{Lung Opacity Detection (LOD)} uses chest X-ray images to classify lung opacity based on data from the RSNA Pneumonia Detection Challenge 2018~\cite{rsna}, annotated by medical practitioners.

\noindent \textbf{ECG Abnormal Detection (EAD)} is an unimodal binary classification task that determines abnormal patterns in 10-second, 12-lead ECG signals from PTB-XL database~\cite{wagner2020ptb}.

\noindent \textbf{Mortality Prediction (MP)} predicts ICU patient survival or death using multimodal dynamic features vital signs, lab tests, input and output, with data sourced from MIMIC-III~\cite{johnson2016mimic}.

\section{Details of Validation Tasks}
\label{sec:validation_tasks}

\noindent \textbf{Enlarged Cardiomediastinum Detection (ECD)}~\cite{irvin2019chexpert} aims to assess the presence of an enlarged cardiomediastinum using medical images from clinical evaluations. This task measures the model's capability to interpret radiographic data effectively.

\noindent \textbf{Sepsis Prediction (SP)} aims to forecast the likelihood of sepsis during ICU stays, testing the model's ability to understand various clinical features. These features are identical to those used in the mortality prediction task, extracted from the MIMIC-III database through the preprocessing pipeline~\cite{van2023yet}.

\noindent \textbf{Medical Visual Question Answering on RAD (MR)} involves using both visual images and textual questions as inputs to generate answers. This task evaluates the model's ability to align text and image modalities within the medical domain. The VQA-RAD dataset is utilized for this task~\cite{lau2018dataset}.

\noindent \textbf{Signal Noise Clarification (SNC)} is a generative task that focuses on accurately describing noise in ECG signals based on corresponding textual questions. The data is extracted from an existing ECG question-answering dataset~\cite{oh2024ecg}. The signals are recorded in 12 channels and last for 10 seconds, similar to the ECG Abnormal Detection task.

\noindent \textbf{Pleural Effusion Detection (PED)}~\cite{irvin2019chexpert} is derived from the CheXpert dataset and involves using X-ray images to identify the presence of pleural effusion, testing the model's ability to interpret radiographic data.

\noindent \textbf{Atelectasis Detection (AD)}~\cite{irvin2019chexpert} also uses X-ray images from the CheXpert dataset to detect atelectasis, evaluating the model's capability in analyzing medical images.

\noindent \textbf{Medical Visual Question Answering on Slake (MS)}~\cite{liu2021slake} utilizes both visual images and textual questions from the SLAKE dataset to generate answers, assessing the model's proficiency in aligning text and image modalities in the medical domain.

\noindent \textbf{Ectopic Beats Detection (EBD)} aims to identify ectopic beats in ECG signals sourced from the PTB-XL database~\cite{wagner2020ptb}.

\section{Dataset Details}
\label{sec:datasets}

For tasks involved in training, we adopt the data partition setup detailed in Section~\ref{sec:data}. For tasks utilized in zero-shot evaluation, we select a subset of corresponding datasets to facilitate the inference efficiency. We cover 1,000 randomly sampled samples for the tasks of Sepsis Prediction, MedVQA-Slake, MedVQA-RAD, Signal Noise Clarification. For Ectopic Beats Detection, we cover 1,000 positive cases and 1,000 negative cases randomly sampled from the dataset, as the original annotations concerning ectopic beats are highly sparse. For Enlarged Cardiomediastinum Detection, Pleural Effusion Detection and Atelectasis Detection, we leverage an existing validation set, which involves 234 samples. Statistics about the datasets leveraged in this study are available in Table~\ref{tab:datasets}. 

\section{Modality Encoding}
\label{sec:encoding}
We list all encoders along with corresponding modalities in Table~\ref{tab:encoders}. Note that all these encoders can be flexibly replaced with other qualified encoders under the framework of {\model}.

\begin{table}[!h]
    \centering
    \caption{Details of modality-specific encoders.}
    \begin{tabular}{c|c}
    \hline
      \textbf{Modality}   &  \textbf{Encoder} \\\hline
      Image   & Deit-tiny~\cite{touvron2021training} \\\hline
      Signal   & CNN~\cite{lecun1998gradient} \\\hline
      Vital Sign   &  Transformer~\cite{vaswani2017attention}  \\\hline
      Lab Results   & Transformer~\cite{vaswani2017attention}  \\\hline
      Input   & Transformer~\cite{vaswani2017attention}\\\hline
      Output  & Transformer~\cite{vaswani2017attention}\\\hline
    \end{tabular}
    \label{tab:encoders}
\end{table}

\section{Single-task Fine-tuning Evaluation}

In addition to the evaluation discussed in Section~\ref{sec:evaluation}, another auxiliary topic worth investigating is whether {\model} can enhance a model's performance on a single task. This is particularly meaningful for practitioners who aim to address a specific task, given the scarcity and specialization often associated with medical data. Therefore, we designed a single-task fine-tuning evaluation, where {\model} is applied to the foundation model for each individual training task. The experimental results are presented in Table~\ref{tab:single_task}.

The results verify that {\model}, benefiting from a well-designed knowledge injection strategy, outperforms both baselines in most tasks. This exploration demonstrates the applicability of {\model} even when tasks for injection are limited, thereby broadening its application scope in complex medical scenarios.